%% file: main.tex
\pdfoutput=1

\documentclass[11pt]{article}

\usepackage{acl}

\usepackage[T1]{fontenc}

\usepackage[utf8]{inputenc}

\usepackage{microtype}

\usepackage{tabu} 
\usepackage{bbm}
\usepackage{transparent}

\usepackage{caption}
\usepackage{multirow}
\usepackage{CJKutf8}
\usepackage{nccmath}

\usepackage{balance}
\usepackage{pifont}

\usepackage{times}
\usepackage{latexsym}
\usepackage{tikz}

\usepackage{booktabs}
\usepackage{diagbox,tabularx}
\usepackage{graphicx}
\usepackage{subfig}
\usepackage{xspace}

\usepackage{fontawesome}
\usepackage{bm}
\usepackage{dsfont}
\usepackage{amsmath}
\usepackage{amssymb}
\usepackage{amsthm}
\usepackage{algorithm}
\input{defs}

\title{Data-Driven Adaptive Simultaneous Machine Translation}

\author{Guangxu Xun$^1$, Mingbo Ma$^1$, Yuchen Bian$^1$, Xingyu Cai$^1$, Jiaji Huang$^1$, \\
\textbf{Renjie Zheng$^1$, Junkun Chen$^2$, Jiahong Yuan$^1$, Kenneth Church$^1$, Liang Huang$^1$} \\
  $^1$Baidu Research, Sunnyvale, CA, USA \\
  $^2$Oregon State University, Corvallis, OR, USA \\
  \texttt{guangxuxun@baidu.com} \\}
  
\begin{document}
\maketitle
\begin{abstract}
In simultaneous translation (SimulMT), the most widely used strategy is the \waitk policy 
thanks to its simplicity and effectiveness in balancing translation quality and latency.
However, \waitk suffers from two major limitations: (a) it is a fixed policy that can not adaptively adjust  latency given context,
and (b) its training is much slower than full-sentence translation.
To alleviate these issues, we propose a novel and efficient training scheme for adaptive SimulMT 
by
augmenting the training corpus with adaptive prefix-to-prefix pairs, 
while the training complexity remains the same as that of training full-sentence translation models. Experiments on two language pairs show that our method outperforms all strong baselines in terms of translation quality and latency.

\end{abstract}

\section{Introduction}
\input{intro}

\section{Adaptive Prefix Pair Generation}
\input{generation}

\section{Data-driven Adaptive Prefix Training}
\input{training}

\section{Experiments}
\input{exp}


\section{Conclusions}
\input{conclusions}

\bibliography{main}
\bibliographystyle{acl_natbib}

\newpage
\appendix
\input{appendix}

\end{document}

%% file: defs.tex
\newcommand{\notes}[1]{}

\theoremstyle{definition}

\theoremstyle{plain}

\newcommand{\ith}[1]{\ensuremath{i^{{th}}}}

\newcount\permx
\newcount\permy
\def\permdot#1#2{
\permx=#1 \advance\permx by-1
\permy=#2 \advance\permy by-1
\psframe[fillcolor=black, fillstyle=solid]
(\permx,\permy)(#1, #2)
}

\newcommand{\boxnum}[1]{{\setlength{\fboxsep}{1pt}\raisebox{1pt}{\hspace{1pt}\fbox{\tiny #1}\hspace{1pt}}}}
\newcommand{\ind}[1]{\ensuremath{_{\kern-0.5pt\boxnum{#1}}}}

\newcommand{\smallnt}[1]{\ensuremath{_{\mbox{\tiny PP}}}\xspace}

\newcommand{\pseudocode}{Algorithm}
\floatname{algorithm}{\pseudocode}

\iffalse

\else

\fi

\newcommand{\waitk}{\ensuremath{\text{wait-$k$}}\xspace}
\newcommand{\Waitk}{\ensuremath{\text{Wait-$k$}}\xspace}

\usepackage[algo2e,ruled,vlined,lined,noend]{algorithm2e}


%% file: intro.tex
Simultaneous Machine Translation (SimulMT) which starts translation before the source sentence finishes, is broadly used in many scenarios which require much shorter translation latency for spontaneous communication such as international conferences, traveling and negotiations.
Most existing SimulMT models \cite{ma+:2019,ma2019monotonic,arivazhagan2019monotonic,arivazhagan2020re,ma-2020-simulmt,ren-2020-simulspeech}
exploit the \waitk policy~\cite{ma+:2019} as a fundamental design of their framework due to the simplicity and effectiveness of the \waitk policy. 

\begin{figure}[t!]
\centering
\includegraphics[width=0.8\linewidth]{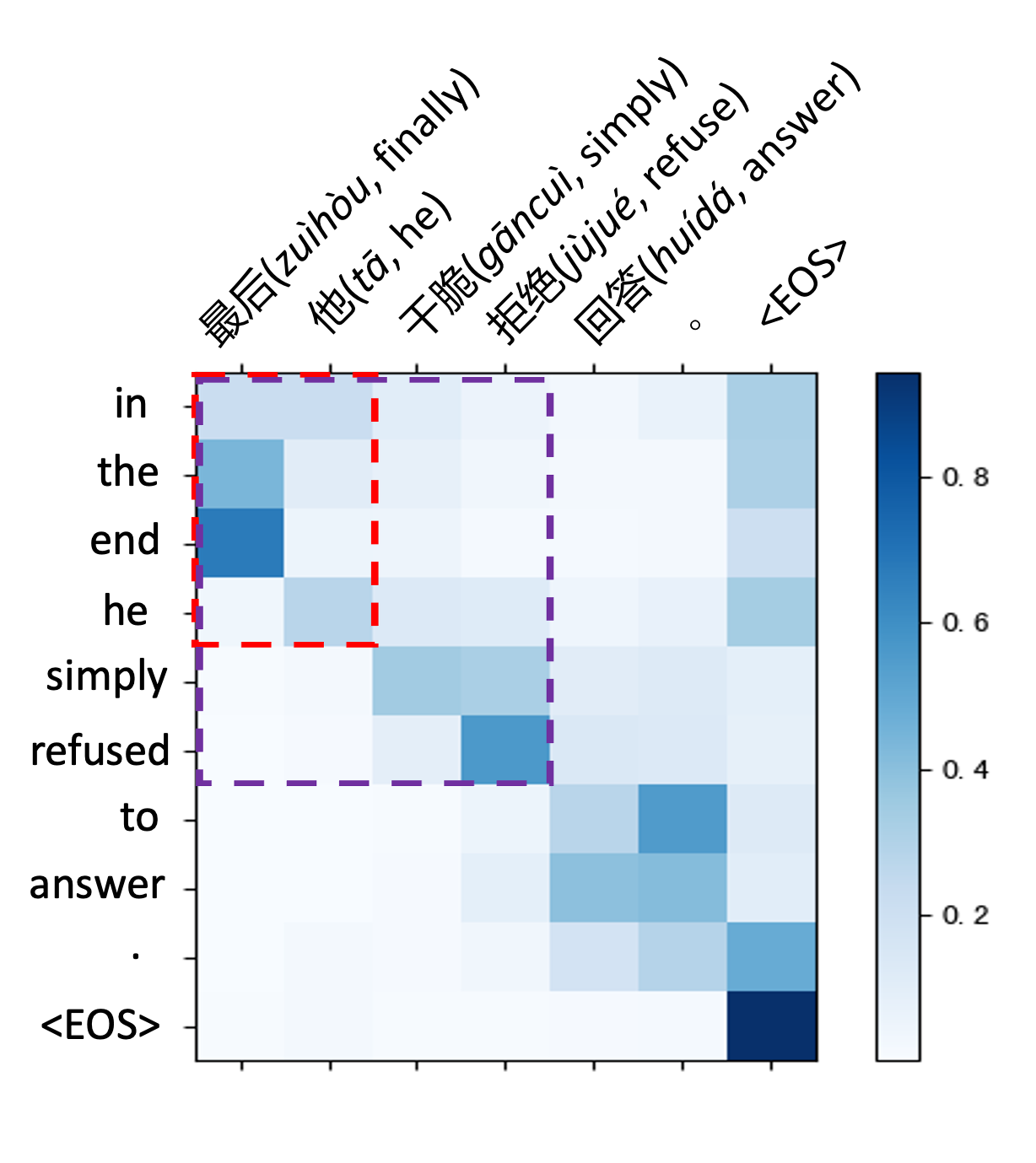}
\caption{Attention map generated by AdaData for a Zh$\rightarrow$En full-sentence pair. The dashed boxes illustrate two pairs of self-translatable prefixes. Given a source prefix, a target prefix is translatable if the dashed attention block contains enough information for translation.}
\label{fig:attnmap}
\end{figure}

However, there are two major drawbacks of \waitk policy. 
Firstly, the \waitk policy 
is computationally expensive and requires substantial GPU memory. 
Given a certain value of $k$, the \waitk policy iterates over all possible 
prefix pairs between source and target sentences to enforce the model to learn 
to anticipate future information of the source sentence. 
This process escalates the original time complexity from quadratic in 
full-sentence translation to cubic in SimulMT.
Secondly, the \waitk policy is a fixed decoding strategy which can not adjust the latency
given difference circumstances.
Some researchers adapted the \waitk policy into 
adaptive policies~\cite{zheng+:2019, zheng2020simultaneous}.
But those methods need to train multiple \waitk models with different $k$, 
which escalate the training complexity even further.



To perform adaptive SimulMT training without escalating the training
time and computation resource, we
propose a data-driven adaptive SimulMT system, named AdaData. 
Our AdaData system automatically generates self-translatable prefix pairs. 
We define a self-translatable prefix pair if the source prefix 
reaches a certain information threshold to obtain 
the target translation prefix. 
Then those generated prefix pairs enable us to train an adaptive SimulMT model in the fashion of full-sentence training. 
The advantages are twofold: first, the SimulMT knowledge resides in the training data instead of in the 
model and hence not necessary for model modifications; second, by sampling 
over the generated prefix pairs, the training complexity remains the same as 
full-sentence translation training. Figure \ref{fig:attnmap} illustrates how to generate two
pairs of self-translatable prefixes based on the source-target attention map. Moreover, the 
generated data allows us to fine-tune a pre-trained full-sentence model 
and dramatically improves the training efficiency. Experiments on De$\rightarrow$En 
and Zh$\rightarrow$En simultaneous translation show substantial improvements 
for both AdaData and AdaData fine-tuning over strong baselines.




%% file: generation.tex
SimulMT consists of two subtasks: the partial translation task when the source sentence is incomplete and the full-sentence translation task when the source sentence is fully observed. This inspires us to treat SimulMT as a multi-task learning problem, where the learned model is able to perform both partial and full-sentence translation depending on the completeness of the source sentence.

In order to learn partial translation, we need to generate artificial prefix pairs from full-sentence pairs to train the model. The major challenge in prefix pair generation is to determine the lengths of the source prefix and the target prefix. As both translation quality and latency are required by SimulMT systems, the generated prefix pairs should meet two basic rules. First, the source prefix should carry necessary information to generate the target prefix for translation quality. Second, the target prefix should contain as many words as possible to minimize translation latency.

To fulfill the two requirements, one key point is to measure the information carried by each source word to predict a target word. The idea of weighted information is similar to the attention mechanism where each word is assigned an attention weight. Therefore, we utilize the Long Short-Term Memory (LSTM) \cite{hochreiter1997long} architecture with the attention mechanism \cite{bahdanau+:2014} to help us generate prefix pairs. In this way, the cumulative information of a source prefix to predict a specific target word can be measured by the sum of its attention weights. In order to obtain accurate cumulative information of a source prefix, we propose the Monotonic LSTM model (MonoLSTM). Compared with the regular attention LSTM model, MonoLSTM utilizes a unidirectional encoder and cut off the connection between the last encoder hidden state and the initial decoder hidden state. By doing so, MonoLSTM prevents source word information from flowing backward and from directly flowing into the decoder. Consequently, the attention module is the only information pathway from the source side to the target side, enabling us to generate self-translatable prefix pairs based on the attention weights.

Specifically, given a full sentence pair of a source sentence $\{x_{1:S}\}$ and a target sentence $\{y_{1:T}\}$, MonoLSTM can generate a $T \times S$ attention matrix $\mathbb{A}$ where each attention score $\alpha_{ts}$ is only affected by prefixes $\{x_{1:s}\}$ and $\{y_{1:t}\}$. For more details about MonoLSTM, refer to Appendix. So the cumulative information a source prefix $\{x_{1:s}\}$ provides to generate the target word $y_t$ can be measured by the sum of attention weights:

\begin{equation}
    \sigma_{t, 1:s} = \sum_{i=1}^s \alpha_{ti}.
\label{eq:cumu}
\end{equation}

Given a hyperparameter threshold $e$, we mathematically define that a target prefix $\{y_{1:t}\}$ is translatable from a prefix $\{x_{1:s}\}$ if all words in the target prefix are translatable from the source prefix, i.e., $\forall j \leq t,$ we have $\sigma_{j,1:s} \geq e$. Hence, we can generate adaptive prefix pairs based on this cumulative information. The data generation process is illustrated in Algorithm \ref{alg:generation}. In addition, we can also adjust the latency trade-off by changing the threshold $e$.

\begin{algorithm2e}[!t]
\SetInd{0.1em}{1em}
\SetAlgoVlined
\KwIn{Full-sentence pairs, threshold $e$ and MonoLSTM}
\KwOut{Adaptive prefix pairs}
\SetAlgoLined
\For{\text{each full pair $(\{x_{1:S}\}, \{y_{1:T}\})$}}
{
    generate source-target attention matrix $\mathbb{A}$ using MonoLSTM\;
    \For{$s\in\{1,...,S\}$}
    {
        \For{$t\in\{1,...,T\}$}
        {
            \uIf{$\sigma_{t, 1:s} < e$}
            {
            break\;
            }
        }
        add $(\{x_{1:s}\}, \{y_{1:t-1}\})$ to prefix pairs\;
    }
}
\Return prefix pairs
\caption{Prefix pair generation}
\label{alg:generation}
\end{algorithm2e}

%% file: training.tex
\subsection{Adaptive Training}
In order to balance the performance on both partial translation and full-sentence translation, we subsample the generated prefix pairs and train the model on a 1:1 mix of prefix pairs and full-sentence pairs for each training epoch. Since prefix pairs are shorter than full-sentence pairs, the multi-tasking training is more efficient than training two full-sentence translation models.

\subsection{Adaptive Inference}
Since the simultaneous translation model is jointly trained for partial translation and full-sentence translation, it is able to adaptively generate the translation at each decoding step. To be more specific, when a new source word arrives, the system starts to decode until the `<eos>' token is generated. This adaptive inference not only closes the gap between training and inference, but also allows the model to adaptively tune the simultaneous translation latency based on the source sequence.

\subsection{Adaptive Fine-tuning}
Another advantage of data-driven adaptive training is that we can significantly improve the training efficiency by fine-tuning a full-sentence translation model using the generated adaptive prefix pairs. In our experiments, we found that 1 epoch of fine-tuning on the 1:1 mixed data is able to achieve comparable performances as the models trained from scratch.

%% file: exp.tex
\subsection{Datasets and Evaluation Metrics}
We conduct experiments on the WMT15 German-to-English dataset (De $\rightarrow$ En, 4.5M sentence pairs) and the NIST Chinese-to-English dataset (Zh $\rightarrow$ En, 2M sentence pairs). For De $\rightarrow$ En, we use newstest-2013 for validation and use newstest-2015 for test. For Zh $\rightarrow$ En, we use NIST 2006 for validation and use NIST 2008 for test. Both datasets are tokenized using BPE \cite{sennrich+:2015}.

We use BLEU \cite{BLEU:2002} to evaluate the translation quality and use Average Lagging (AL) \cite{ma+:2019} to measure the translation latency. Since each Chinese sentence in the Zh $\rightarrow$ En dataset has 4 English references, we report 4-reference BLEU scores for this dataset.

\subsection{Baselines and Model Configurations}
The baseline models we include in the experiments are \Waitk \cite{ma+:2019}, MILk \cite{arivazhagan2019monotonic}, MMA-H, MMA-IL \cite{ma2019monotonic}, Proportional re-translation \cite{niehues2018low, arivazhagan2020re}, and a policy-based adaptive model \cite{zheng2020simultaneous}.

To be consistent with the baseline models, we adopt the Transformer-base \cite{vaswani+:2017} architecture as our simultaneous translation model. It consists of a 6-layer encoder and a 6-layer decoder. The model dimension is set to 512 and the number of attention heads is 8. Since the MMA-H model and the MMA-IL model are based on 1024-dimensional Transformers, we also carry out experiments with a wider configuration of our model to make a fair comparison with MMA-H and MMA-IL. Specifically, as with MMA-H and MMA-IL, the wider version of our model has 1024 dimensions and the number of attention heads is set 16. The number of layers remains to be 6.

For the adaptive prefix generation part, our proposed MonoLSTM is based on a 2-layer, 512-dimensional unidirectional LSTM. Please find more training details in Appendix.

\subsection{Training Efficiency}

\begin{figure}[t!]
\centering
\includegraphics[width=1\linewidth]{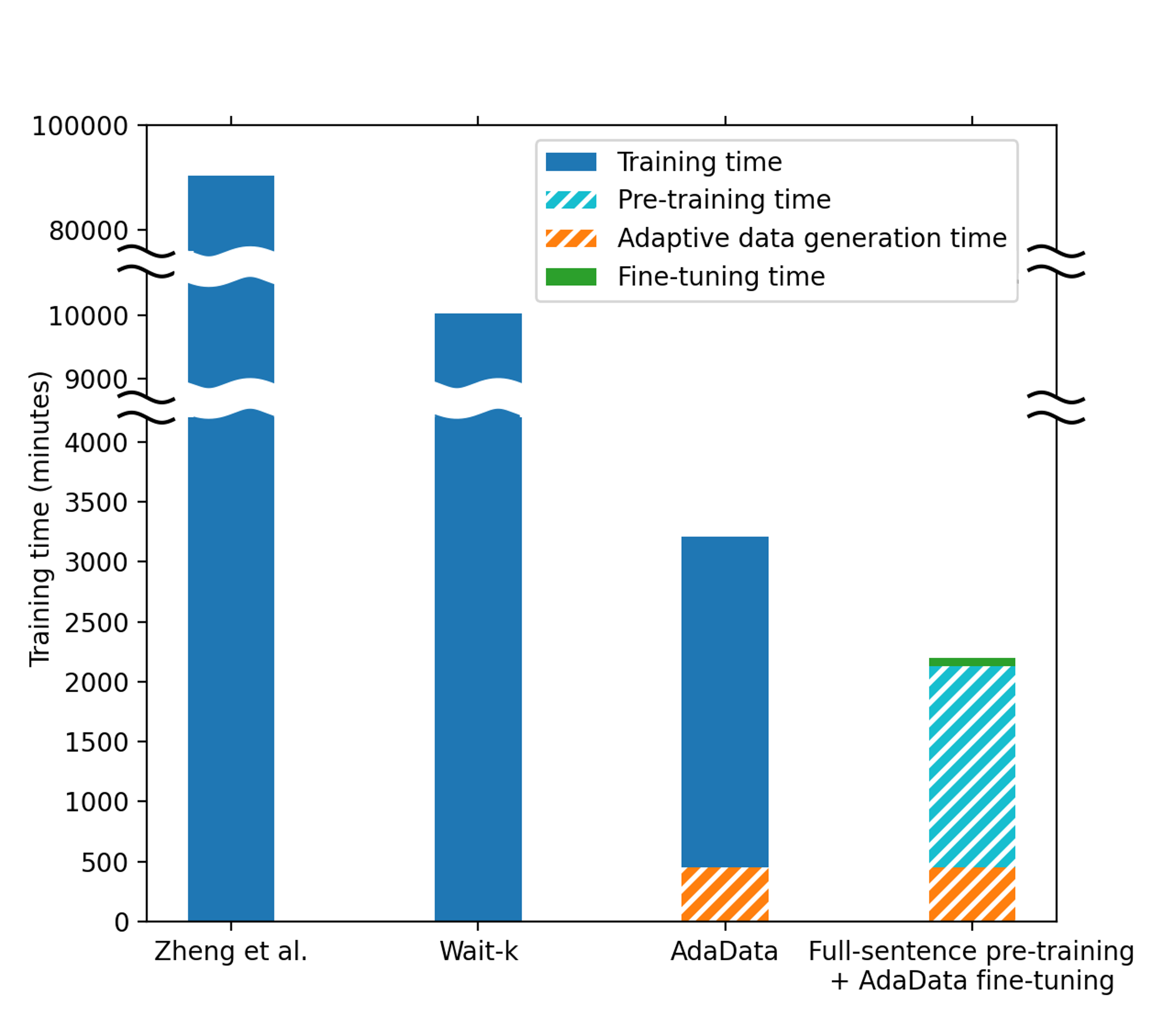}
\caption{Training efficiency comparison of \Waitk, AdaData and AdaData fine-tuning on the De$\rightarrow$En dataset using 10 $\times$ 1080Ti GPUs.}
\label{fig:trainingtime}
\end{figure}

Figure \ref{fig:trainingtime} illustrates the training efficiency of \Waitk, \cite{zheng2020simultaneous} and our methods. Note that, the hatched bars represent ``run only once". For example, we can pre-train a full-sentence model once and fine-tune multiple times with different information threshold $e$ for different quality-latency trade-offs. The fine-tuned models are only fine-tuned for 1 epoch. We can see that our methods achieve significantly better training efficiency, especially AdaData fine-tuning.

\begin{figure}[!t]
\centering
\includegraphics[width=1\linewidth]{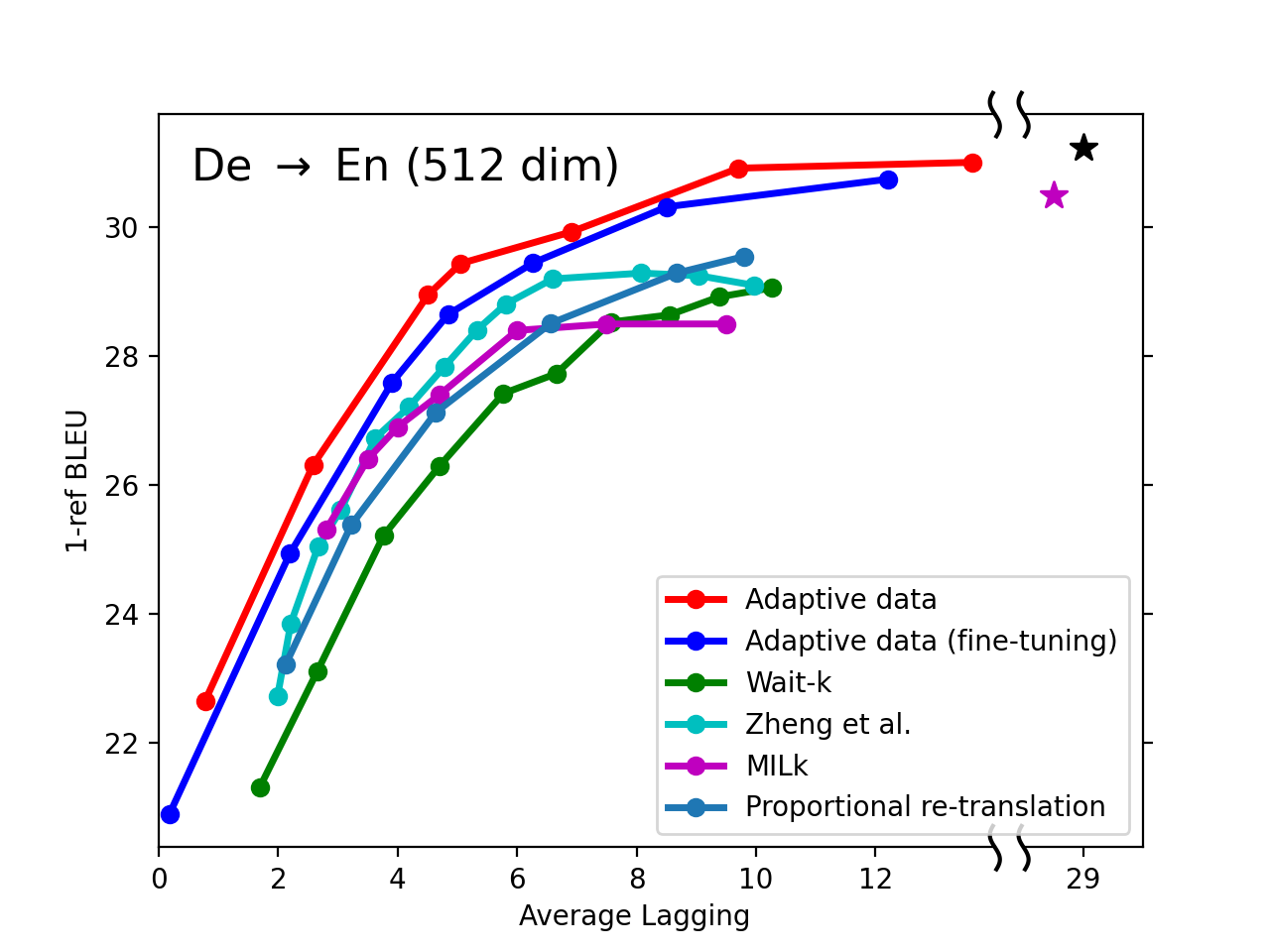}
\caption{Translation quality against latency curves for De $\rightarrow$ En test sets with the basic 512 dimension configuration.}
\label{fig:de-en-512}
\end{figure}

\begin{figure}[!t]
\centering
\includegraphics[width=1\linewidth]{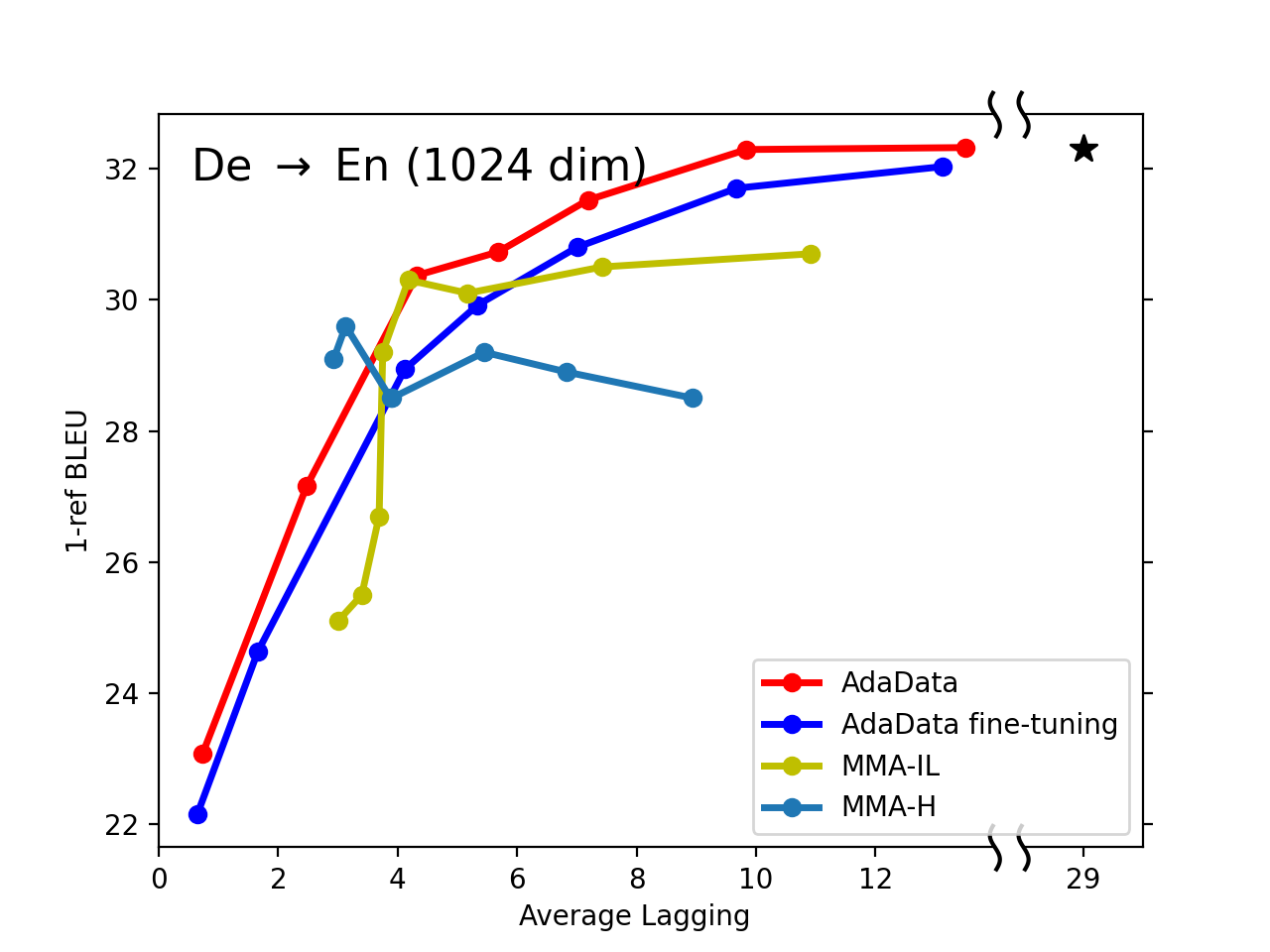}
\caption{Translation quality against latency curves for De $\rightarrow$ En test sets with the wide 1024 dimension configuration.}
\label{fig:de-en-1024}
\end{figure}

\subsection{Performance Comparison}
In order for our model to obtain different quality-latency trade-offs, we set the cumulative information threshold $e$ ranging from 0.1 to 0.7 with a step size of 0.1. The 512-dimensional model comparison results and the 1024-dimensional model comparison results on the De $\rightarrow$ En test sets are reported in Figure \ref{fig:de-en-512} and Figure \ref{fig:de-en-1024} respectively. MILk is based on LSTMs, resulting in a lower full-sentence BLEU score compared with other Transformer based models. The erasure rate of Proportional re-translation is set to 0 to match other models. Proportional re-translation is also a data-driven method. In their papers, they discussed two ways to generate prefix pairs: proportion and word alignments. They also found that both methods achieve similar performances. So we report the proportion method here for simplicity. Since neither ways can accurately measure the cumulative information in prefixes, their performances are worse than ours. To sum up, compared with the baseline models, our proposed model achieves better quality-latency trade-offs and larger areas under curves.

\begin{figure}[!t]
\centering
\includegraphics[width=1\linewidth]{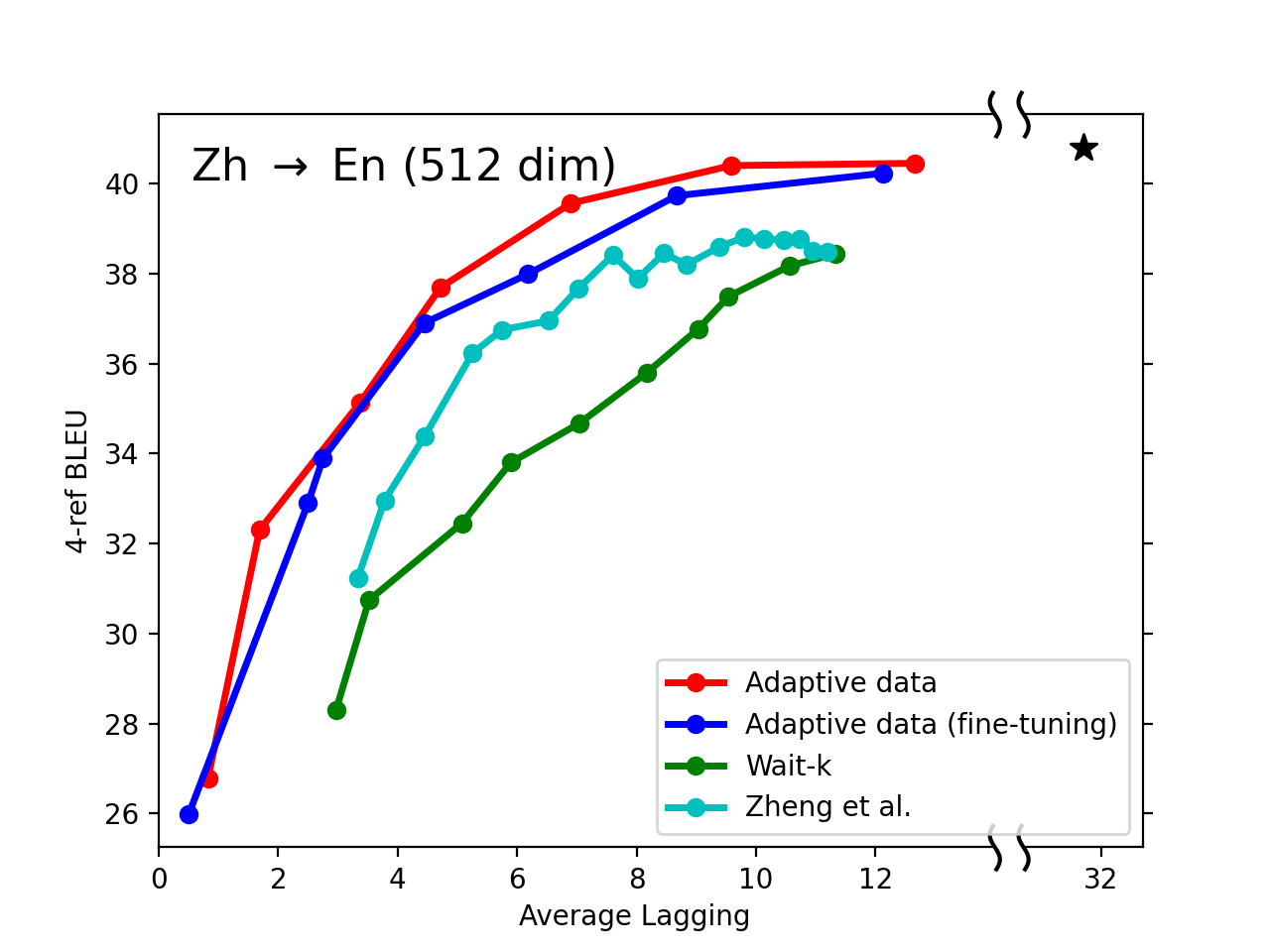}
\caption{Translation quality against latency curves for Zh $\rightarrow$ En test sets with the basic 512 dimension configuration.}
\label{fig:zh-en}
\end{figure}

The comparison results on the Zh $\rightarrow$ En test sets are presented in Figure \ref{fig:zh-en}. We can see that our proposed model is still able to outperform the baseline models.

It is noteworthy that our fine-tuned models are also able to achieve better performances than the baseline models. Considering the models are fine-tuned for only 1 epoch, this demonstrates that the generated adaptive data can help us greatly improve the training efficiency without significantly hurting the model performance.

%% file: conclusions.tex
We have designed an efficient data-driven algorithm for adaptive SimulMT. The proposed method is able to measure the information requirement for each decoding step and generate self-translatable prefix pairs for partial translation. Experiments validate the effectiveness of the proposed data-driven adaptive SimulMT system. Moreover, the generated prefix pairs enable us to fine-tune pre-trained full-sentence translation models and dramatically improve the training efficiency.

%% file: appendix.tex
\section{Details of MonoLSTM}
\begin{figure}
\centering
\includegraphics[width=0.47\textwidth]{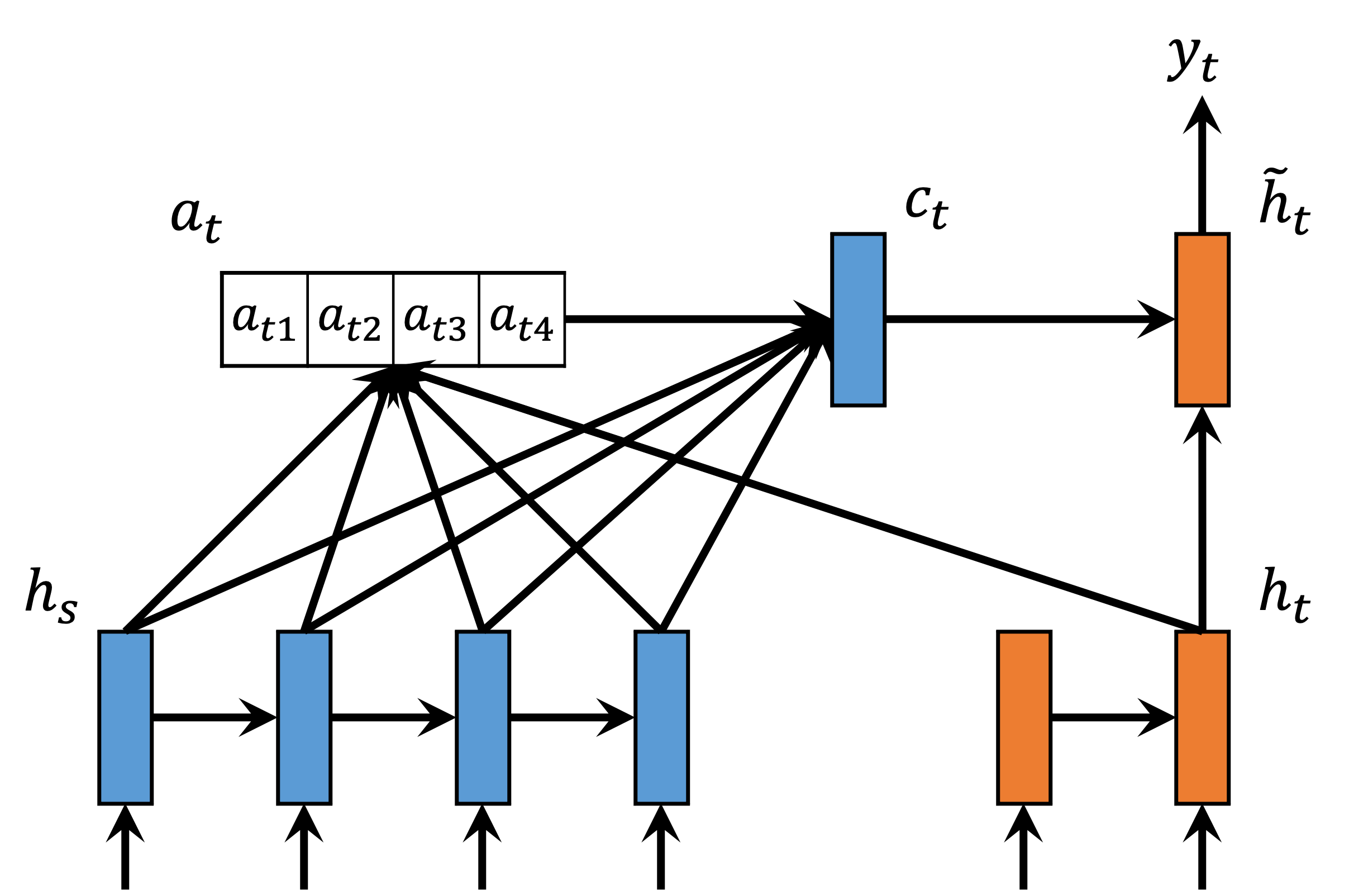}
\caption{MonoLSTM.}
\label{fig:monolstm}
\end{figure}

The framework of MonoLSTM is shown in Figure \ref{fig:monolstm}. Compared with the regular attention LSTM model, MonoLSTM utilizes a unidirectional encoder and cut off the connection between the last encoder hidden state and the initial decoder hidden state. By doing so, MonoLSTM prevents source word information from flowing backward and from directly flowing into the decoder. Thus, the attention module is the only information pathway from the source side to the target side, ensuring that at each decoding step, the attention score of each source word is only affected by the current word and the previous words.

More specifically, given a full sentence pair of a source sentence $\{x_{1:S}\}$ and a target sentence $\{y_{1:T}\}$, at each decoding time step $t$, target word $y_t$ is predicted as:

\begin{equation}
    p(y_t|y_{<t}, x) = \text{softmax}(\boldsymbol{W} \tilde{\boldsymbol{h}}_t),
\label{eq:yt}
\end{equation}
where $\tilde{\boldsymbol{h}}_t$ is the concatenation of the target hidden state $\boldsymbol{h}_t$ and the source-side context vector $\boldsymbol{c}_t$. The context vector $\boldsymbol{c}_t$ represents the summary of the source sentence at time step $t$, which is computed as the weighted sum of the source hidden states based on their attention scores:

\begin{equation}
    \boldsymbol{c}_t = \sum_s \alpha_{ts} \bar{\boldsymbol{h}}_s,
\label{eq:ct}
\end{equation}
where $\bar{\boldsymbol{h}}_s$ is the source hidden state and $\alpha_{ts}$ denotes the attention score between target word $y_t$ and source word $x_s$:

\begin{equation}
    \alpha_{ts} = \frac{\exp (\boldsymbol{h}_t^\intercal \bar{\boldsymbol{h}}_s)}{\sum_{i=1}^S \exp (\boldsymbol{h}_t^\intercal \bar{\boldsymbol{h}}_i)}.
\label{eq:ats}
\end{equation}

The attention scores are normalized and sum up to 1. We can see that, in MonoLSTM $\bar{\boldsymbol{h}}_s$ does not contain information for the source words after $x_s$ by removing the backward LSTM and $\boldsymbol{h}_t$ does not contain information for the source sentence by cutting off the connection between source hidden states and target hidden initialization. Therefore, the unnormalized attention score, i.e., the numerator of $\alpha_{ts}$ is only affected by the source prefix $\{x_{1:s}\}$ and the target prefix $\{y_{1:t}\}$. Then we can use the attention scores to measure the information a source prefix provides to produce a target prefix.

\section{Details of Training Configurations}
For the 512-dimensional models on De $\rightarrow$ En, they are trained on 10 $\times$ 1080Ti GPUs. It consists of a 6-layer encoder and a 6-layer decoder. The model dimension is set to 512, the intermediate dimension is 2048 and the number of attention heads is 8. The number of max tokens is 6144. The optimizer is Adam with betas set to (0.9, 0.98). The learning rate is 0.00015 and the warm-up steps are 2000. We also reduce the learning rate on plateaus with a patience of 4. Weight decay is 0.000001. Label smoothing is 0.1 and dropout rate is 0.3.

For the 1024-dimensional models on De $\rightarrow$ En, they are trained on 8 $\times$ TitanX GPUs. It consists of a 6-layer encoder and a 6-layer decoder. The model dimension is set to 1024, the intermediate dimension is 4096 and the number of attention heads is 16. The training configurations follow \cite{ma2019monotonic}. The number of max tokens is 3584. The optimizer is Adam with betas set to (0.9, 0.98). The learning rate is 0.0005 and the warm-up steps are 6000. The learning rate scheduler is inverse\_sqrt. The warm-up initial learning rate is 1e-7. Label smoothing is 0.1 and dropout rate is 0.3.

For the 512-dimensional models on Zh $\rightarrow$ En, they are trained on 10 $\times$ 1080Ti GPUs. It consists of a 6-layer encoder and a 6-layer decoder. The model dimension is set to 512, the intermediate dimension is 2048 and the number of attention heads is 8. The number of max tokens is 10240. The optimizer is Adam with betas set to (0.9, 0.98). The learning rate is 0.00025 and the warm-up steps are 500. We also reduce the learning rate on plateaus with a patience of 4. Weight decay is 0.000001. Label smoothing is 0.1 and dropout rate is 0.3.

For the adaptive prefix generation part, our proposed MonoLSTM is based on a 2-layer, 512-dimensional unidirectional LSTM. It is also trained on 10 $\times$ 1080Ti GPUs. The number of max tokens is 20480. The optimizer is Adam with betas set to (0.9, 0.997) and epsilon set to 1e-9. The learning rate is 0.005 and the warm-up steps are 4000. We also reduce the learning rate on plateaus with a patience of 4. Weight decay is 0.000001. Label smoothing is 0.1 and dropout rate is 0.2.

For De $\rightarrow$ En, the BPE vocabulary for De and En are 16k and 16k, respectively. For Zh $\rightarrow$ En, the BPE vocabulary for Zh and En are 20k and 10k, respectively.

For fine-tuning, all training configuration remain the same except that learning rate is multiplied by 0.1.


%% file: main.bbl
\begin{thebibliography}{14}
\expandafter\ifx\csname natexlab\endcsname\relax\def\natexlab#1{#1}\fi

\bibitem[{Arivazhagan et~al.(2019)Arivazhagan, Cherry, Macherey, Chiu, Yavuz,
  Pang, Li, and Raffel}]{arivazhagan2019monotonic}
Naveen Arivazhagan, Colin Cherry, Wolfgang Macherey, Chung-Cheng Chiu, Semih
  Yavuz, Ruoming Pang, Wei Li, and Colin Raffel. 2019.
\newblock Monotonic infinite lookback attention for simultaneous machine
  translation.
\newblock \emph{arXiv preprint arXiv:1906.05218}.

\bibitem[{Arivazhagan et~al.(2020)Arivazhagan, Cherry, Macherey, and
  Foster}]{arivazhagan2020re}
Naveen Arivazhagan, Colin Cherry, Wolfgang Macherey, and George Foster. 2020.
\newblock Re-translation versus streaming for simultaneous translation.
\newblock \emph{arXiv preprint arXiv:2004.03643}.

\bibitem[{Bahdanau et~al.(2014)Bahdanau, Cho, and Bengio}]{bahdanau+:2014}
Dzmitry Bahdanau, Kyunghyun Cho, and Yoshua Bengio. 2014.
\newblock Neural machine translation by jointly learning to align and
  translate.
\newblock \emph{arXiv preprint arXiv:1409.0473}.

\bibitem[{Hochreiter and Schmidhuber(1997)}]{hochreiter1997long}
Sepp Hochreiter and J{\"u}rgen Schmidhuber. 1997.
\newblock Long short-term memory.
\newblock \emph{Neural computation}, 9(8):1735--1780.

\bibitem[{Ma et~al.(2019{\natexlab{a}})Ma, Huang, Xiong, Zheng, Liu, Zheng,
  Zhang, He, Liu, Li, Wu, and Wang}]{ma+:2019}
Mingbo Ma, Liang Huang, Hao Xiong, Renjie Zheng, Kaibo Liu, Baigong Zheng,
  Chuanqiang Zhang, Zhongjun He, Hairong Liu, Xing Li, Hua Wu, and Haifeng
  Wang. 2019{\natexlab{a}}.
\newblock \href {https://doi.org/10.18653/v1/P19-1289} {{STACL}: Simultaneous
  translation with implicit anticipation and controllable latency using
  prefix-to-prefix framework}.
\newblock In \emph{Proceedings of the 57th Annual Meeting of the Association
  for Computational Linguistics}, pages 3025--3036, Florence, Italy.
  Association for Computational Linguistics.

\bibitem[{Ma et~al.(2019{\natexlab{b}})Ma, Pino, Cross, Puzon, and
  Gu}]{ma2019monotonic}
Xutai Ma, Juan Pino, James Cross, Liezl Puzon, and Jiatao Gu.
  2019{\natexlab{b}}.
\newblock Monotonic multihead attention.
\newblock \emph{arXiv preprint arXiv:1909.12406}.

\bibitem[{Ma et~al.(2020)Ma, Pino, and Koehn}]{ma-2020-simulmt}
Xutai Ma, Juan Pino, and Philipp Koehn. 2020.
\newblock {S}imul{MT} to {S}imul{ST}: Adapting simultaneous text translation to
  end-to-end simultaneous speech translation.
\newblock In \emph{Proceedings of the 1st Conference of the Asia-Pacific
  Chapter of the Association for Computational Linguistics and the 10th
  International Joint Conference on Natural Language Processing}.

\bibitem[{Niehues et~al.(2018)Niehues, Pham, Ha, Sperber, and
  Waibel}]{niehues2018low}
Jan Niehues, Ngoc-Quan Pham, Thanh-Le Ha, Matthias Sperber, and Alex Waibel.
  2018.
\newblock Low-latency neural speech translation.
\newblock \emph{arXiv preprint arXiv:1808.00491}.

\bibitem[{Papineni et~al.(2002)Papineni, Roukos, Ward, and Zhu}]{BLEU:2002}
Kishore Papineni, Salim Roukos, Todd Ward, and Wei-Jing Zhu. 2002.
\newblock Bleu: a method for automatic evaluation of machine translation.
\newblock In \emph{Proceedings of ACL}, pages 311--318, Philadephia, USA.

\bibitem[{Ren et~al.(2020)Ren, Liu, Tan, Zhang, Qin, Zhao, and
  Liu}]{ren-2020-simulspeech}
Yi~Ren, Jinglin Liu, Xu~Tan, Chen Zhang, Tao Qin, Zhou Zhao, and Tie-Yan Liu.
  2020.
\newblock {S}imul{S}peech: End-to-end simultaneous speech to text translation.
\newblock In \emph{Proceedings of the 58th Annual Meeting of the Association
  for Computational Linguistics}.

\bibitem[{Sennrich et~al.(2015)Sennrich, Haddow, and Birch}]{sennrich+:2015}
Rico Sennrich, Barry Haddow, and Alexandra Birch. 2015.
\newblock Neural machine translation of rare words with subword units.
\newblock \emph{arXiv preprint arXiv:1508.07909}.

\bibitem[{Vaswani et~al.(2017)Vaswani, Shazeer, Parmar, Uszkoreit, Jones,
  Gomez, Kaiser, and Polosukhin}]{vaswani+:2017}
Ashish Vaswani, Noam Shazeer, Niki Parmar, Jakob Uszkoreit, Llion Jones,
  Aidan~N Gomez, \L{}ukasz Kaiser, and Illia Polosukhin. 2017.
\newblock Attention is all you need.
\newblock In \emph{Advances in Neural Information Processing Systems 30}.

\bibitem[{Zheng et~al.(2020)Zheng, Liu, Zheng, Ma, Liu, and
  Huang}]{zheng2020simultaneous}
Baigong Zheng, Kaibo Liu, Renjie Zheng, Mingbo Ma, Hairong Liu, and Liang
  Huang. 2020.
\newblock Simultaneous translation policies: From fixed to adaptive.
\newblock In \emph{Proceedings of the 58th Annual Meeting of the Association
  for Computational Linguistics}.

\bibitem[{Zheng et~al.(2019)Zheng, Zheng, Ma, and Huang}]{zheng+:2019}
Baigong Zheng, Renjie Zheng, Mingbo Ma, and Liang Huang. 2019.
\newblock Simultaneous translation with flexible policy via restricted
  imitation learning.
\newblock In \emph{ACL}.

\end{thebibliography}
